\documentclass{interact}

\usepackage[T1]{fontenc}
\usepackage[utf8]{inputenc}

\usepackage{epstopdf}
\usepackage[caption=false]{subfig}

\usepackage[numbers,sort&compress]{natbib}
\bibpunct[, ]{[}{]}{,}{n}{,}{,}

\makeatletter
\def\NAT@def@citea{\def\@citea{\NAT@separator}}
\makeatother
\usepackage{amsmath,amssymb}
\usepackage{graphicx}
\graphicspath{{./}}
\usepackage{microtype}
\usepackage[hidelinks]{hyperref}

\newcommand{\PHITSBench}{PHITSBench}
\newcommand{\CMS}{\mathit{CMS}}
\newcommand{\EX}{\mathit{EX}}
\newcommand{\PF}{\mathit{PF}}

\begin{document}

\articletype{Article}

\title{\PHITSBench: an execution-scored benchmark for AI-assisted PHITS radiation-transport input generation using natural language}

\author{
\name{Xianglin Ji and Svetlana V. Boriskina\thanks{CONTACT Svetlana Boriskina. Email: sborisk@mit.edu}}
\affil{Department of Mechanical Engineering, Massachusetts Institute of Technology, Cambridge, MA 02139, USA}
}

\maketitle

\begin{abstract}
We introduce \PHITSBench, an execution-scored benchmark for the Monte Carlo Particle and Heavy Ion Transport code System (PHITS). \PHITSBench{}  comprises 282 transport-scorable tasks spanning three common workflow categories: parameter editing (\emph{Edit}), syntax repair (\emph{Repair}), and complete simulation generation from natural-language descriptions (\emph{Reproduce}). Each task is evaluated using a Composite Metric Score that combines execution success with agreement between generated and reference transport observables. Using \PHITSBench, we evaluate five GPT-5.4-based configurations ranging from zero-shot prompting to knowledge-augmented and agentic workflows. Without domain-specific knowledge, the model performs well on editing and repair tasks (95\% and 70\% success, respectively) but fails to generate correct simulations from scratch (0\% success on the \emph{Reproduce} track). A structured, machine-readable PHITS knowledge catalog, supplied alongside the user manual, raises single-shot \emph{Reproduce}-task success to 57\%. Agentic execution provides a further improvement to 66-73\%, but at increased computational cost. Failure analysis shows that the remaining errors are dominated by incorrect selection and configuration of physical observables rather than syntax generation. These results suggest that future progress in AI-assisted radiation-transport modeling will depend as much on machine-readable knowledge bases, curated domain-training datasets, and execution-grounded evaluation environments as on advances in foundation models themselves.

\end{abstract}

\begin{keywords}
PHITS; Monte Carlo radiation transport; large language models; benchmark; input file generation; agentic systems
\end{keywords}

\section{Introduction}
\label{sec:intro}

Accurate modeling of radiation transport is essential for nuclear energy and fusion engineering, particle accelerator operation, medical physics, outer space exploration, and advanced manufacturing, which often require analysis and design of systems operating in radiation-rich environments. Monte Carlo (MC) radiation transport codes such as PHITS \citep{sato2024}, MCNP \citep{werner2018}, GEANT4 \citep{agostinelli2003}, and FLUKA \citep{ferrari2005} are state-of-the-art tools for predicting how neutrons, photons, electrons, and ions interact with materials in arbitrary three-dimensional (3D) geometries. These platforms encode decades of validated physics, but their simulation setup, execution, and validation of results remain challenging even for experienced designers. Constructing a valid simulation requires extensive domain expertise in radiation physics, geometry modeling, materials, source characterization, and code-specific input languages. In practice, a significant fraction of the effort associated with a radiation transport study lies not in executing simulations, but in constructing, debugging, validating, and maintaining complex simulation input decks. This creates a substantial barrier to productivity, limits rapid exploration of large design spaces, and restricts access to advanced modeling capabilities outside a relatively small community of experts.

\begin{figure}[t]
\centering
\includegraphics[width=\linewidth]{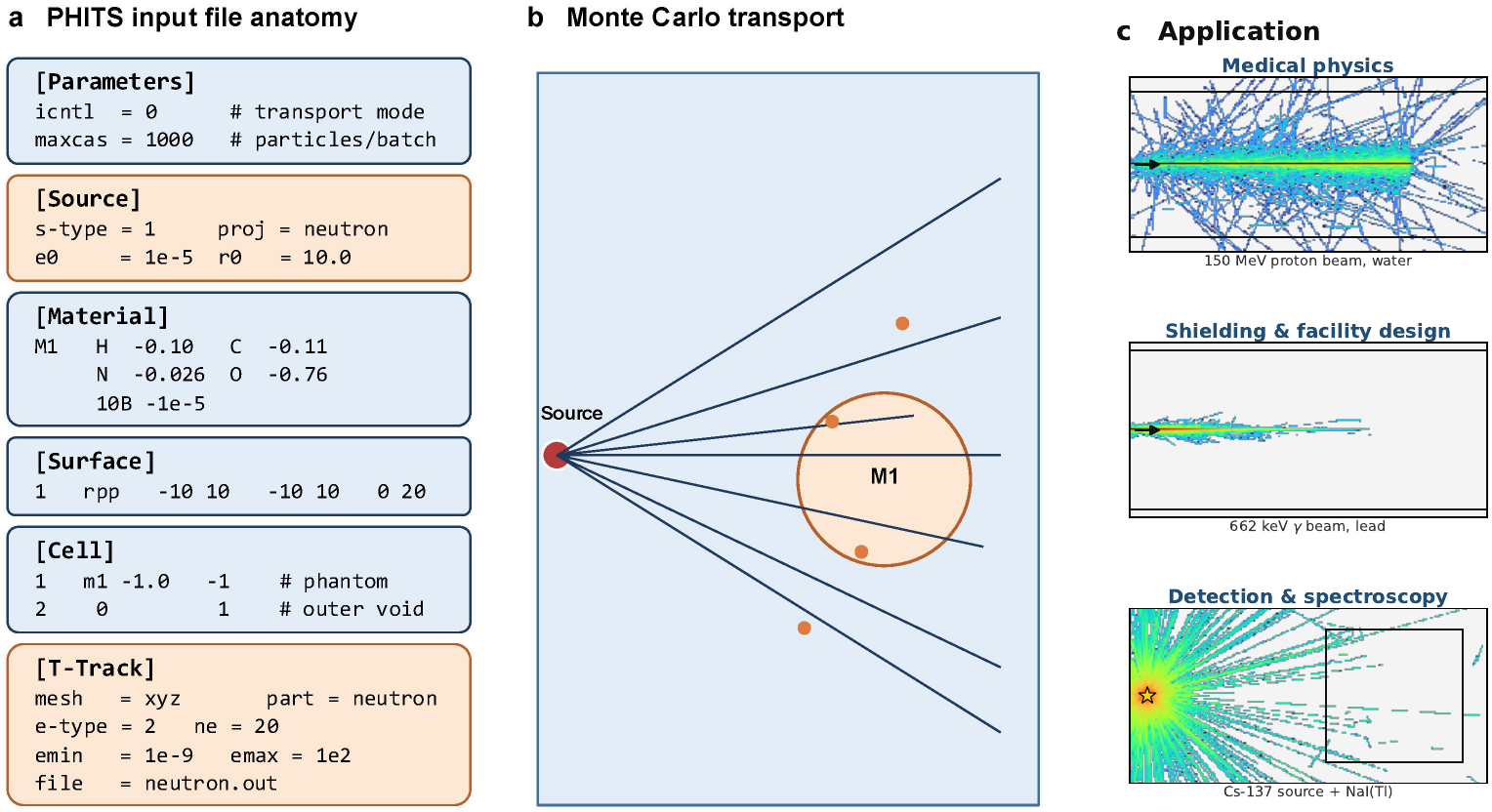}
\caption{PHITS at a glance. (a) A PHITS input file consists of a sequence of named keyword sections defining simulation parameters, particle sources, materials, geometry, and tally requests. (b) During execution, particles are emitted from the specified source, transported through the geometry, and scored according to user-defined \texttt{[T-*]} tally definitions. In the example shown, a neutron source emits neutrons that propagate through the geometry and interact with the material region M1 (orange), producing secondary particles at the interaction points (orange dots); the simulation returns the neutron track-length flux scored by a \texttt{[T-Track]} tally.  (c) The same radiation-transport framework underpins diverse application areas; the panels show three representative PHITS simulations from this study---medical physics (a proton beam in water), shielding and facility design (a photon beam in lead), and radiation detection and spectroscopy (a Cs-137 source and a NaI(Tl) detector).}
\label{fig:phits-overview}
\end{figure}

Recent advances in large language models (LLMs) and agentic AI systems have created new opportunities to automate scientific computing workflows. LLMs now achieve excellent performance on many conventional code-generation benchmarks and increasingly demonstrate utility in engineering design, scientific programming, and computational modeling \citep{chen2021, jimenez2024, lai_2023_ds1000, liu2023, zhuo_2025_bigcodebench, tian2024}. A growing body of work has begun to extend these capabilities to scientific simulation environments. In radiation transport, AutoFLUKA demonstrated AI-assisted generation of MC simulations through domain-knowledge-augmented agents \citep{ndum2025}. In finite-element analysis, MCP-SIM introduced a self-correcting multi-agent framework for FEniCS-based simulations  \citep{park2026}, while Shafiq et al. evaluated LLM performance on Gmsh and Elmer workflows for geometry generation and solver setup \citep{shafiq2025}. In computational fluid dynamics, OpenFOAMGPT employed retrieval-augmented generation for OpenFOAM-based simulations \citep{pandey_2025_openfoamgpt}, and MetaOpenFOAM developed a specialized multi-agent framework for CFD workflows \citep{chen_2024_metaopenfoam}, motivating the creation of the CFDLLMBench benchmark suite \citep{somasekharan_2025_cfdllmbench}. Similar efforts have recently emerged for general multiphysics simulation, including MooseAgent for MOOSE-based workflows  \citep{zhang2025mooseagent}, and for nuclear engineering applications, including AutoSAM for generation of input files for the SAM reactor thermal-hydraulic code \citep{abulawi2026autosam}. Related benchmark efforts have also appeared in adjacent simulation domains, including network simulation \citep{ahmed2025simcode}, molecular dynamics \citep{holbrook2026lammps}, particle-physics experiment design and simulation  \citep{grace2026}, and nuclear energy applications \citep{ornl2025}. 

LLM performance for scientific computing is often limited by access to domain-specific knowledge rather than by reasoning capability. To address this challenge, recent studies have shown that retrieval-augmented generation (RAG) and direct documentation injection can substantially improve code generation and simulation setup by providing structured access to technical information that is either absent from pretraining data or difficult to retrieve from long documents \citep{pandey_2025_openfoamgpt, zhou2023docprompting, pimparkhede2024doccgen}. In parallel, agentic workflows have emerged as a complementary strategy when static knowledge alone is insufficient. Modern multi-agent systems decompose complex tasks into specialized roles, including planning, execution, verification, and repair, as demonstrated by frameworks such as MetaGPT \citep{hong_2024_metagpt}, AutoGen \citep{wu_2023_autogen}, SciAgents \citep{ghafarollahi_2024_sciagents}, and SWE-agent \citep{yang_2024_sweagent}. These systems build upon iterative reasoning and self-correction paradigms such as ReAct \citep{yao2023}, Self-Refine \citep{madaan_2023_selfrefine}, and Reflexion \citep{shinn2023}, enabling models to incorporate execution feedback and progressively improve generated outputs. However, other studies suggest that self-correction is not a universal solution. Huang et al. \citep{huang2024} and Olausson et al. \citep{olausson2024} showed that the effectiveness of self-repair is fundamentally bounded by the quality and informativeness of external feedback, while Kamoi et al. \citep{kamoi2024} provided a comprehensive survey of the limitations and opportunities of self-correcting LLM systems. 

Overall, there is an emerging consensus in the scientific computing community that LLMs can assist with simulation setup and execution, particularly when augmented with retrieval mechanisms, structured documentation, and iterative self-correction. However, existing efforts share several important limitations. Most target software ecosystems are extensively documented online, are represented in public training corpora, and rely on modern scripting languages and APIs. Thus, it remains difficult to determine whether the observed performance reflects genuine reasoning about physical systems or the retrieval of previously seen examples \citep{jain_2024_livecodebench}. Furthermore, many existing benchmarks focus primarily on syntactic correctness, while placing less emphasis on the physical validity of results.

Radiation transport simulations provide a compelling testbed for evaluating AI-assisted scientific computing. Successful generation of a valid simulation requires simultaneous reasoning about geometry, materials, source definitions, physics models, tally specifications, and code-specific syntax. Small errors may lead to failed execution, silently incorrect outputs, or physically invalid results. As a result, radiation transport workflows offer a realistic measure of whether AI systems can bridge the gap between natural-language scientific intent and executable, physically meaningful simulations. 

The Particle and Heavy Ion Transport code System (PHITS) presents an especially challenging and informative case study. PHITS is one of the most widely used general-purpose MC packages, which is routinely applied to reactor shielding analysis, accelerator design, detector modeling, proton therapy, medical physics, and space-radiation assessment. Unlike many modern software environments, PHITS relies on a specialized input language derived from legacy scientific computing workflows; its bracketed sections with column-sensitive lines and ad-hoc subsection grammar resemble 1980s Fortran-card-deck input. Furthermore, the code, documentation, and example libraries are not broadly available on the public web, making PHITS underrepresented in the data used to train contemporary foundation models. 

Importantly, PHITS input generation requires more than code synthesis. A valid simulation combines multiple interdependent components, including geometry definitions, material specifications, source descriptions, transport parameters, and tally requests.  Input is specified in a free-format text file consisting of named keyword \emph{sections}, including [Parameters], [Source], [Material], [Surface], [Cell], [T-Track], [T-Deposit], etc. (Fig.~\ref{fig:phits-overview}). Although the language is relatively permissive in formatting, it is unforgiving in physics. A geometry whose outer void fails to enclose the simulation region can trigger a fatal error; a tally configured with an incorrect energy-bin definition may execute successfully while producing physically meaningless results; and keywords that appear similar across sections may carry different numerical meanings. Consequently, generating PHITS inputs requires simultaneous understanding of both radiation physics and parser-specific constraints.

To systematically evaluate the capability of LLMs to generate such inputs, we introduce PHITSBench, an execution-scored benchmark for AI-assisted PHITS input generation. PHITSBench contains 282 transport-scorable tasks spanning three categories that mirror common PHITS workflows: parameter editing, syntax repair, and complete simulation reconstruction from natural-language descriptions. Unlike benchmarks scored by text-similarity metrics such as CodeBLEU~\citep{ren2020codebleu}, PHITSBench evaluates generated inputs by executing PHITS simulations and comparing the resulting transport observables against reference solutions. This execution-linked framework enables direct assessment of both syntactic correctness and physical fidelity.

Using PHITSBench, we investigate three central questions. First, to what extent can LLMs generate valid radiation transport simulations without domain-specific knowledge? Second, how much performance can be recovered through structured knowledge injection and retrieval mechanisms? Third, can specialized agentic architectures outperform generic coding agents in the generation and repair of complex scientific simulations?  The answers provide insight not only into the future of AI-assisted radiation transport modeling, but more broadly into the role of agentic AI systems in scientific computing, nuclear engineering, and physics-based design.

\section{Materials and methods}
\label{sec:matmeth}

\subsection{Benchmark design}
\label{sec:bench}

\begin{figure}[t]
\centering
\includegraphics[width=\linewidth]{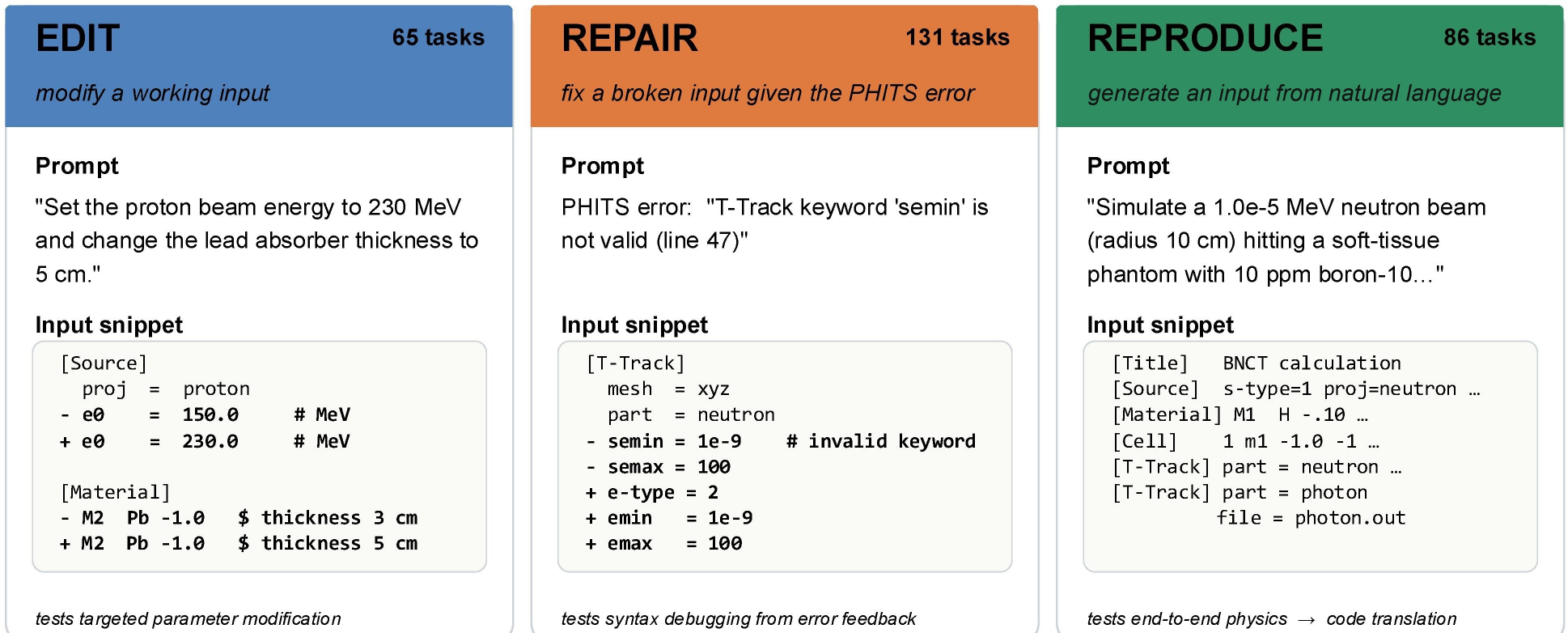}
\caption{Representative examples of tasks in each benchmark track. \emph{Edit} (left): the model receives a valid PHITS input file together with a natural-language modification request and must generate an updated input whose transport results match the modified reference simulation. \emph{Repair} (center): the model receives a deliberately corrupted PHITS input file and the corresponding PHITS error message and must generate a corrected simulation input. \emph{Reproduce} (right): the model receives only a natural-language description of the physical problem and must generate a complete PHITS input file from scratch.}
\label{fig:task-examples}
\end{figure}

\PHITSBench{} was developed to evaluate AI-assisted generation of radiation transport simulations across the spectrum of common PHITS user workflows. The benchmark contains 282 tasks derived from the official PHITS 3.34 example library, which are organized into three task categories: \emph{Edit, Repair} and \emph{Reproduce} (Fig.~\ref{fig:task-examples}).

The \emph{Edit} track includes 65 tasks ($n = 65$) and evaluates controlled modification of existing reference simulations. A model receives a valid PHITS input file together with a natural-language instruction, which requests one or more changes to the simulation setup, such as modifying source energy, material composition, geometry dimensions, or tally parameters. Hand-crafted modification instructions range from a single parameter change to composite edits touching source energy, material
composition, and geometry simultaneously. The generated output is expected to reproduce the behavior of a reference simulation, in which the same modifications were applied manually. This task reflects routine parameter studies and design-space exploration commonly performed by PHITS users.

The \emph{Repair} track ($n = 131$) evaluates error diagnosis and correction. Each task contains a deliberately corrupted PHITS input file together with the corresponding system-generated error message. Corrupted input files include a single fault at a randomly chosen valid location---a misspelled keyword, a missing required parameter, a corrupted or duplicated section header, or a malformed numeric format---and record the generated PHITS error message. The full taxonomy of edit modifications and repair faults, with task counts per category, is given in Table~\ref{tab:error-types}. The model must identify the source of the error and generate a corrected simulation input. These tasks represent common debugging activities encountered during simulation development and validation.

The \emph{Reproduce} track ($n = 86$)  is the most challenging and is the primary focus of this work. In this setting, the model receives only a natural-language description of the physical problem, including geometry, materials, source conditions, and desired outputs. The model must generate a complete PHITS input file from scratch. Successful completion therefore requires simultaneous reasoning about radiation physics, geometry construction, tally selection, and PHITS syntax.

\begin{table}
\tbl{Taxonomy of the \emph{Edit} (modification) and \emph{Repair} (fault) tracks. Tasks are grouped into thematic categories, and the rightmost column summarizes the corresponding modification or fault represented by each category. Counts ($n$) correspond to the transport-scorable subset used throughout the benchmark.}
{\small
\begin{tabular}{@{}lr p{0.58\textwidth}@{}}
\toprule
Category & $n$ & Sub-types \\
\midrule
\multicolumn{3}{@{}l}{\textbf{Edit track} \quad ($n=65$)} \\
\addlinespace[1pt]
Source                     & 16 & change the beam energy and its energy range \\
\addlinespace[2pt]
Material                   & 20 & replace a material or change its density \\
\addlinespace[2pt]
Geometry                   & 5  & resize or reshape the geometry \\
\addlinespace[2pt]
Tally / mesh               & 1  & change the resolution of the scoring mesh \\
\addlinespace[2pt]
Composite ($\geq 2$ edits) & 23 & two or three of the above edits applied together \\
\midrule
\multicolumn{3}{@{}l}{\textbf{Repair track} \quad ($n=131$)} \\
\addlinespace[1pt]
Section structure & 25 & corrupted, duplicated, or missing section headers \\
\addlinespace[2pt]
Source keywords   & 32 & invalid, misspelled, or missing source parameters \\
\addlinespace[2pt]
Material          & 18 & malformed or missing material, density, or isotope entries \\
\addlinespace[2pt]
Cell / surface    & 21 & bad cell or region syntax, or an undefined surface \\
\addlinespace[2pt]
Tally keywords    & 35 & unknown or missing tally settings, such as the output file, an axis, the mesh type, or the energy bins \\
\bottomrule
\end{tabular}}
\label{tab:error-types}
\end{table}

To ensure that benchmark evaluation reflects physically meaningful transport simulations,  \PHITSBench{} is restricted to tasks that execute in transport mode (icntl = 0) and generate tally outputs suitable for quantitative comparison. Non-transport control modes like all-vacuum mode, 2D geometry plotting mode, and source checking mode were excluded because they do not produce transport observables that can be evaluated consistently. Every task was generated and reviewed by a domain expert. Tasks containing prompt-reference mismatches, spurious whitespace differences, or overly complex geometries were corrected or removed during the revision process. 

\subsection{Composite Metric Score}
\label{sec:cms}

Evaluating AI-generated scientific simulations requires assessing both \emph{syntactic validity} and \emph{physical correctness}. A generated PHITS input may execute successfully while producing physically incorrect outputs, whereas an otherwise valid transport model may fail because of a minor syntax or formatting error. To capture both aspects of performance, PHITSBench employs an execution-grounded metric termed the \emph{Composite Metric Score} (CMS). The CMS combines an \emph{execution score}, $\EX$, and a \emph{physics-fidelity score}, $\PF$:
\begin{equation}
\CMS = w_{\EX}\EX + w_{\PF}\PF ,
\end{equation}
where $\EX \in \{0,1\}$, $\PF \in [0,1]$, and $w_{\EX}$, $w_{\PF}$ are weighting coefficients.

The execution score evaluates whether the generated input produces a valid PHITS simulation. A task receives $\EX=1$ only if PHITS terminates successfully, reports no \texttt{** Error} or \texttt{*** ERROR} entries in \texttt{phits.out}, and generates all expected output files. Otherwise, $\EX=0$. The physics-fidelity score quantifies agreement between the generated and reference transport solutions. For each expected tally output, the corresponding PHITS results are parsed and converted into one-dimensional numerical arrays. Similarity is then evaluated using three complementary measures: (i) normalized $L^2$ distance, (ii) agreement in total integral magnitude, and (iii) agreement in peak position. These metrics capture distribution shape, overall normalization, and key physical features, respectively.

Because different tally outputs contain different amounts of information, the relative weighting of these three components depends on tally resolution. High-resolution distributions provide sufficient information to evaluate shape, whereas low-resolution or integral quantities do not. The weighting scheme used throughout the benchmark is summarized in Table~\ref{tab:cms_weights}. Component weights were selected \emph{a priori} based on the information content of each tally type and were not tuned using benchmark results. For a given task, tally-level fidelity scores are averaged across all expected outputs. If a simulation fails to execute ($\EX=0$), the corresponding physics-fidelity score is defined as $\PF=0$.

\begin{table}
\tbl{Physics-fidelity weighting scheme used in PHITSBench.}
{\begin{tabular*}{\hsize}{@{\extracolsep{\fill}}lccc@{}}
\toprule
Tally type & $L^2$ & Integral & Peak position \\
\midrule
Distribution ($n>20$ bins) & 60\% & 25\% & 15\% \\
Coarse tally ($6 \le n \le 20$) & 40\% & 40\% & 20\% \\
Integral quantity ($n \le 5$) & 0\% & 100\% & 0\% \\
\bottomrule
\end{tabular*}}
\label{tab:cms_weights}
\end{table}

Different benchmark tracks place different emphasis on execution success and physics fidelity. For the \emph{Reproduce} track, generating a runnable simulation from a natural-language description is itself a substantial achievement. We therefore assign equal importance to execution and physics fidelity: $(w_{\EX},w_{\PF})=(0.5,0.5)$. For the \emph{Edit} and \emph{Repair} tracks, where the starting point is already a valid or nearly valid simulation, physics fidelity provides the primary measure of task completion. Accordingly, we use $(w_{\EX},w_{\PF})=(0.2,0.8)$ for \emph{Edit} and $(0.3,0.7)$ for \emph{Repair}.

Throughout this work, a task is considered successfully solved when $\CMS \ge 0.95$. This threshold corresponds to a simulation that both executes correctly and reproduces the reference transport observables with high fidelity. Sensitivity analysis showed that perturbing the track-specific weights by $\pm 0.10$ does not alter the qualitative ranking of methods evaluated in this study, indicating that the reported conclusions are robust to reasonable variations in scoring parameters.
\subsection{Knowledge Engineering for PHITS Simulation}
\label{sec:catalog}

A central objective of this work is to evaluate how different forms of domain knowledge influence AI-assisted generation of PHITS simulations. To this end, we constructed a machine-readable PHITS knowledge catalog and compared multiple knowledge-injection strategies ranging from no external knowledge to structured domain-specific guidance. The experiments reported in this work use three sources of PHITS-related knowledge:

\begin{enumerate}
    \item Model pretraining knowledge, corresponding to whatever PHITS-related information is already present in the foundation model (\textbf{Baseline}).
    \item The PHITS user manual in the .pdf format, which provides comprehensive documentation of simulation syntax, physics models, and example workflows (\textbf{PDF}).
    \item  A structured PHITS knowledge catalog custom-designed for this study (\textbf{Cat}).
\end{enumerate}

The PHITS knowledge catalog was designed to provide a compact, machine-readable representation of information required for simulation generation and validation. Unlike the PHITS manual, which is optimized for human users, the catalog organizes syntax rules, parameter definitions, and parser constraints in a format intended for direct use by LLMs. The catalog was generated through LLM-assisted parsing of PHITS source code and subsequently reviewed, corrected, and expanded by a domain expert. Parser routines responsible for reading and validating PHITS input files were analyzed to recover valid keyword names, accepted value ranges, default values, and inter-keyword dependencies. These rules were then cross-referenced against the PHITS user manual and example library to identify undocumented behaviors and common failure modes. 

The resulting catalog consists of eleven Markdown documents comprising approximately 7,300 lines of text and approximately 69,000 tokens. Each rule is linked to the relevant PHITS input section and, where possible, annotated with source-code provenance for traceability. Importantly, the catalog captures information that is difficult to infer from documentation alone, including parser-specific defaults, context-dependent keyword behavior, and cross-section constraints that frequently cause simulation failures.

\subsection{Evaluation methods}
\label{sec:methods}

To quantify the impact of domain knowledge on simulation generation, we evaluate three progressively richer knowledge configurations.  
\begin{itemize}
    \item The \textbf{Baseline} configuration provides no PHITS-specific information beyond the task description. Performance under this condition reflects the model's existing knowledge acquired during pretraining.
    \item The  \textbf{+PDF} configuration augments the model with access to the complete PHITS user manual. This setup represents a documentation-assisted workflow analogous to retrieval-augmented generation approaches used in other scientific computing domains.
    \item The \textbf{+Cat+PDF} configuration supplements the manual with the structured knowledge catalog. The catalog is provided directly within the model context, while the manual remains available as a reference source. This combination allows the model to access both detailed documentation and a compact representation of parser-level constraints.
    \item The  \textbf{Agentic} methods use the same knowledge sources as the \textbf{+Cat+PDF} configuration, but additionally exploit execution feedback and iterative repair mechanisms.
\end{itemize}

All methods received identical benchmark tasks and used the same underlying foundational model. The primary differences between configurations were (i) the amount of PHITS-specific knowledge available to the model and (ii) whether the model could execute PHITS and iteratively refine its output.

The first three configurations perform a single inference pass and differ only in the knowledge available to the model. The \textbf{Baseline} configuration receives only the benchmark task description and therefore reflects whatever PHITS-related knowledge is already present in the model's pretraining corpus. The \textbf{+PDF} configuration augments the task description with the complete PHITS user manual supplied through the Responses API. The manual remains available throughout the inference process but no execution feedback is provided. The \textbf{+Cat+PDF} configuration additionally supplies the structured PHITS knowledge catalog. The catalog is included directly within the model context, while the full manual remains available as a reference document.

\textbf{Agentic} configurations augment knowledge injection with execution-driven refinement. The first agentic baseline uses \textbf{Codex CLI} \citep{codex_cli}, a general-purpose coding agent capable of executing arbitrary commands and iteratively modifying generated outputs. The agent is provided with the same PHITS manual and knowledge catalog, together with access to a local PHITS installation. The second configuration is a \textbf{Custom Multi-Agent} architecture, which we designed specifically for scientific simulation generation. The multi-agent workflow consists of three role-specialized agents:

\begin{itemize}
    \item \textbf{Composer}: generates an initial PHITS input deck;
    \item \textbf{Manager}: orchestrates execution, verification, and repair;
    \item \textbf{Repairer}: modifies candidate inputs based on execution feedback.
\end{itemize}
The \textbf{Manager} implements a ReAct-style control loop \citep{yao2023}, and coordinates a fixed tool set consisting of:
\begin{itemize}
    \item \verb|compose_phits_input|
    \item \verb|static_check_phits|
    \item \verb|run_phits|
    \item \verb|summarize_failure|
    \item \verb|repair_phits_input|
\end{itemize}

Two additional verification tools provide domain-specific validation:
\begin{itemize}
    \item \verb|inspect_tally_outputs|, which identifies empty or physically implausible tally results;
    \item \verb|check_tally_config|, which compares generated tally configurations against task requirements.
\end{itemize}
The workflow proceeds through iterative cycles of generation, validation, execution, diagnosis, and repair until either a valid solution is obtained or a maximum of five repair iterations is reached. The \textbf{Repairer} additionally has retrieval access to a vector database containing both the PHITS manual and the structured catalog. The implementation uses the OpenAI Agents SDK (v0.8.4) \citep{openai_agents_sdk}.

All experiments used OpenAI GPT-5.4 accessed through the Responses API (April–May 2026 release). Single-shot methods employed medium reasoning effort and a maximum output length of 16,384 tokens. No custom sampling parameters were specified. All reported results correspond to single, non-deterministic runs. The agentic baselines used the same underlying model. Codex CLI (v0.125.0) operated with xhigh reasoning effort and unrestricted iteration. In our \textbf{Custom Multi-Agent} system, the \textbf{Manager} and \textbf{Composer} agents used medium reasoning effort, while the \textbf{Repairer} agent used xhigh reasoning.

Inference cost was estimated from logged token usage using official GPT-5.4 pricing. For the single-shot methods, most of the manual and catalog content is prefix-cached after the first request, reducing subsequent costs. The \textbf{Baseline} configuration costs approximately \$0.01 per task, while \textbf{+PDF} and \textbf{+Cat+PDF} cost approximately \$0.18 and \$0.20 per task, respectively. The unrestricted \textbf{Codex CLI} workflow incurs substantially higher cost because the agent repeatedly accesses documentation and performs multiple repair cycles. Average cost was approximately \$0.82 per task. The bounded \textbf{Custom Multi-Agent} architecture reduces this cost to approximately \$0.13 per task through limited repair iterations, selective use of high-reasoning agents, and retrieval-based access to documentation rather than repeated inclusion in context.

\section{Results}
\label{sec:results}

Table~\ref{tab:per-method}  summarizes benchmark performance across all evaluated methods. We first examine the \emph{Edit} and \emph{Repair} tracks, which approach saturation under single-shot prompting, before turning to the more challenging \emph{Reproduce} track and the effects of knowledge injection and agentic execution. The success rate is defined as the fraction of tasks executed with $\CMS \geq 0.95$.

\begin{table}
\tbl{Performance of the evaluated methods across the three PHITSBench tracks. Entries show the number of tasks achieving $CMS \geq 0.95$ out of the corresponding transport-scorable subset ($n$ shown in the column headers). “n/e” denotes configurations that were not evaluated (see Section~\ref{sec:methods}).}
{\begin{tabular*}{\hsize}{@{\extracolsep{\fill}}lrrr@{}}
\toprule
Method & Edit (n=65) & Repair (n=131) & Reproduce (n=86) \\
\midrule
Baseline & 62 & 92 & 0 \\
+ PDF & 60 & 110 & 28 \\
+ Cat+PDF & 62 & 116 & 49 \\
Codex CLI & n/e & n/e & 63 \\
Our Agent & n/e & n/e & 57 \\
\bottomrule
\end{tabular*}}
\label{tab:per-method}
\end{table}

\subsection{Edit and Repair: single-shot prompting is sufficient}
\label{sec:results-tracks}
\begin{figure}[t]
\centering
\includegraphics[width=\linewidth]{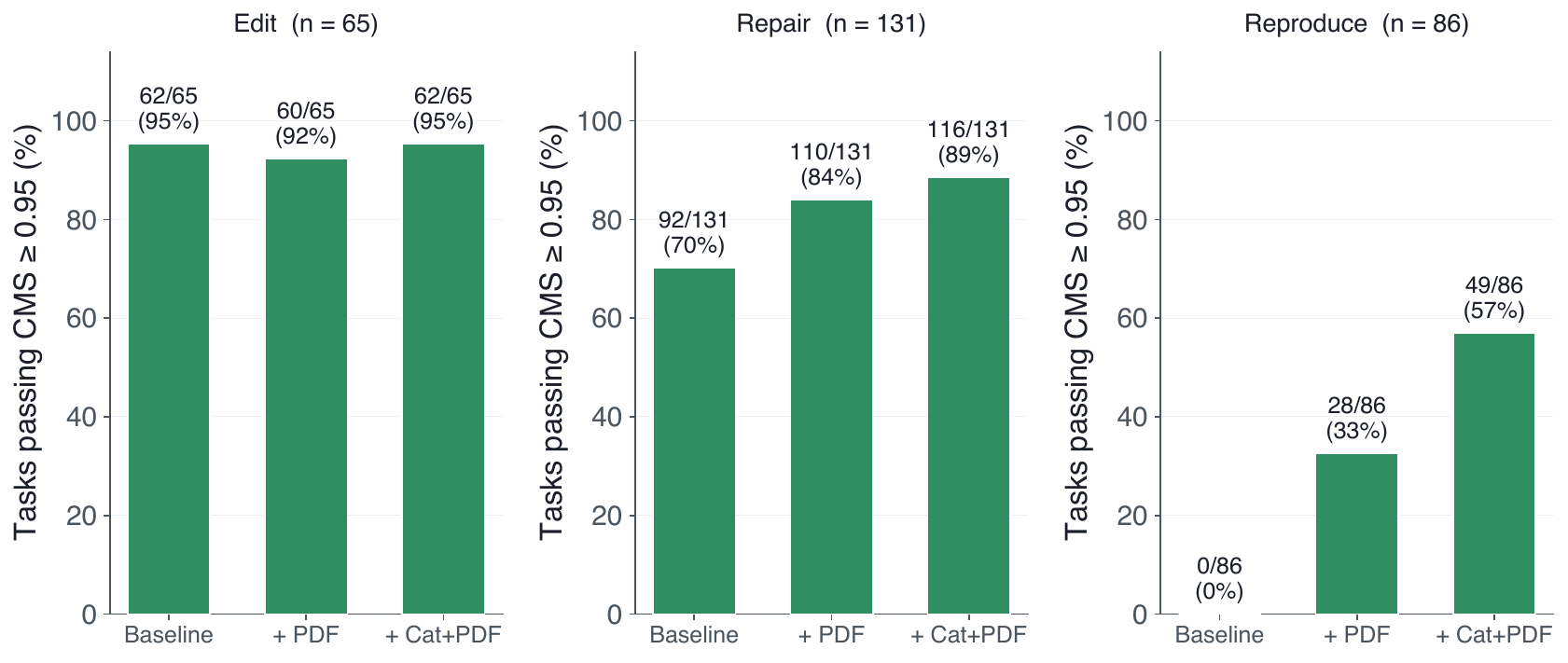}
\caption{Single-shot success rate($\CMS \geq 0.95$)  across all three benchmark tracks for the non-agentic methods. The \emph{Edit} track is effectively saturated even in the zero-shot baseline, and the \emph{Repair} track benefits substantially from progressive knowledge injection (+18\%). In contrast, the same progression improves \emph{Reproduce} performance only to 49 of 86 tasks (57\%), highlighting the remaining gap that motivates the use of agentic methods.}
\label{fig:per-track}
\end{figure}

Figure~\ref{fig:per-track} shows the success rate of the three non-agentic methods across all benchmark tracks. The \emph{Edit} and \emph{Repair} tracks provide the model with substantial contextual information in the form of an existing PHITS input file. Consequently, these tasks primarily evaluate localized modification and debugging rather than full simulation synthesis. 

For the \emph{Edit}  track, performance is effectively saturated even without PHITS-specific knowledge. The \textbf{Baseline} and \textbf{+Cat+PDF} configurations both solve 62 of 65 tasks (95\%), while the \textbf{+PDF} configuration solves 60 of 65 tasks (92\%). The small differences between methods suggest that modifying an existing PHITS input file is largely tractable for a frontier model once a valid simulation structure is already present. Because both \emph{Edit} and \emph{Repair} approaches saturate under single-shot prompting, agentic methods were not evaluated on these tracks. 

The \emph{Repair} track exhibits greater sensitivity to domain knowledge. The \textbf{Baseline} configuration successfully repairs 92 of 131 tasks (70\%), while the addition of the PHITS manual (\textbf{+PDF} ) increases performance to 110 tasks (84\%). Providing both the manual and structured catalog (\textbf{+PDF+Cat}) further improves performance to 116 tasks (89\%). This improvement is expected because \emph{Repair} tasks provide an explicit PHITS error message that localizes the fault. The knowledge sources then supply the information needed to identify the correct keyword, parameter, or syntax structure.

\subsection{Reproduce: from natural language to executable simulation}
\label{sec:results-headline}

\begin{figure}[t]
\centering
\includegraphics[width=\linewidth]{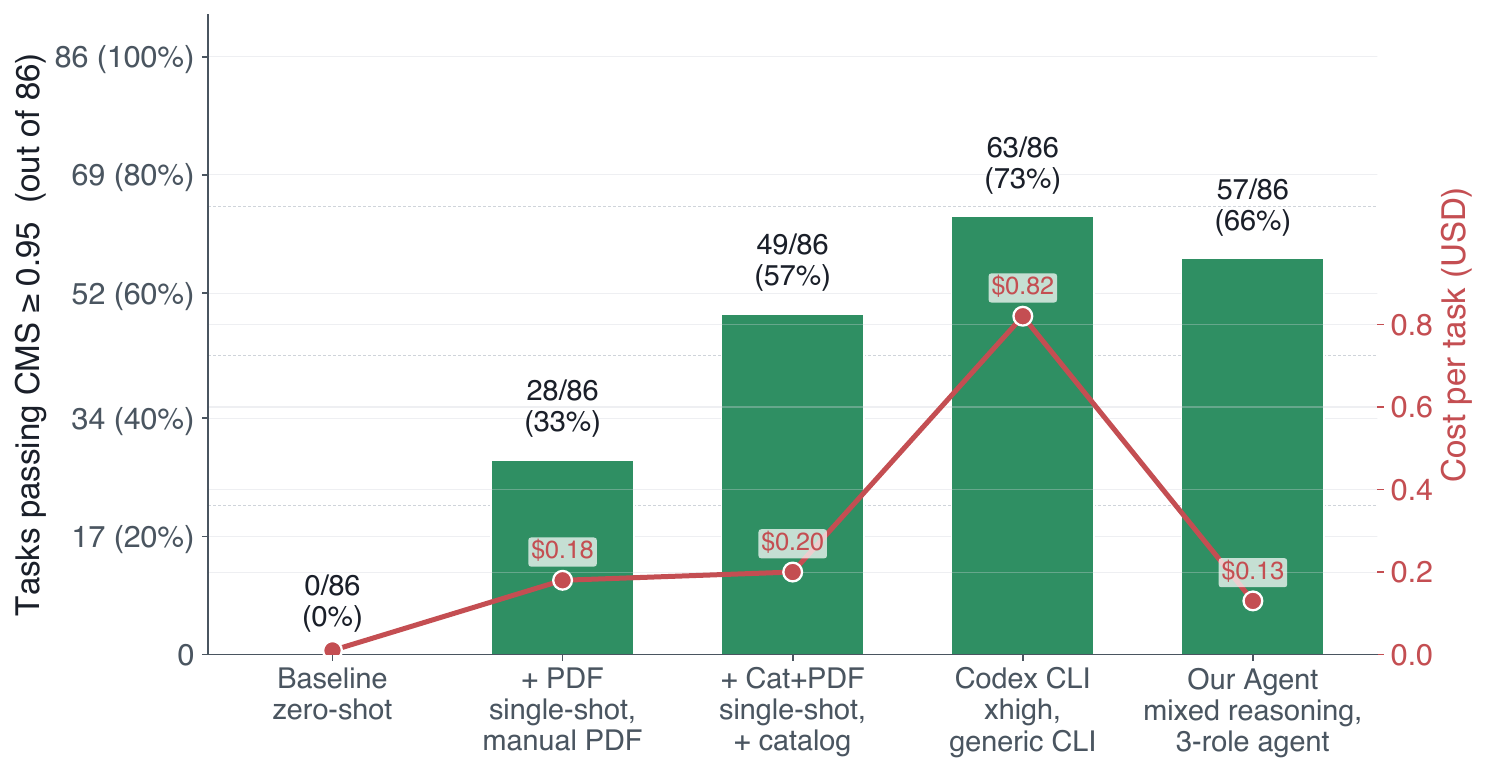}
\caption{Success rate ($\CMS \geq 0.95$) on the 86 \emph{Reproduce}-track tasks across the five evaluated methods (green bars, left axis). Structured knowledge injection (\textbf{+Cat+PDF} relative to the \textbf{Baseline}) accounts for the largest overall performance improvement, while \textbf{Agentic}  execution provides a further increase in success rate. The red line (right axis) shows the estimated per-task inference cost at GPT-5.4 list pricing (single-run estimate; single-shot costs are amortized post-cache values). \textbf{Codex CLI} achieves the highest success rate (73\%) but at approximately six times the per-task cost of our \textbf{Custom Multi-Agent}, which achieves comparable performance (66\%) at substantially lower cost.}
\label{fig:headline}
\end{figure}

The \emph{Reproduce} track represents the most demanding benchmark scenario. Unlike the \emph{Edit} and \emph{Repair} tasks, the model receives neither a valid PHITS template nor an error message. Instead, it must construct an entire simulation from a natural-language description of geometry, materials, particle sources, and desired observables. Figure ~\ref{fig:headline} summarizes performance on the 86 \emph{Reproduce} tasks (see also Table~\ref{tab:per-method}).

The \textbf{Baseline} configuration fails all 86 tasks at the CMS $\geq 0.95$ threshold. Although generated outputs often contain plausible PHITS-like structures, they typically fail because of missing keywords, malformed geometry definitions, inconsistent tally specifications, or other parser-level errors. These results indicate that pretraining alone does not provide sufficient PHITS-specific knowledge to generate valid simulations.

Providing access to the PHITS user manual substantially improves performance. The \textbf{+PDF} configuration successfully solves 28 of 86 tasks (33\%), demonstrating that the model can effectively utilize technical documentation during generation. However, the improvement remains limited despite the manual containing all information required to construct valid simulations.
The greatest gain is obtained through the injection of the structured knowledge. Adding the PHITS catalog (\textbf{+Cat+PDF}) increases performance from 28 to 49 tasks (57\%), corresponding to an additional 21 successful simulations. The catalog reorganizes information contained in the manual into a machine-readable form and explicitly encodes parser constraints, default values, and cross-section dependencies. This result suggests that accessibility and organization of domain knowledge are at least as important as the volume of available documentation.

The \textbf{+Cat+PDF} configuration achieves an average physics-fidelity score of approximately 0.67 across all \emph{Reproduce} tasks. Among the 37 unsuccessful cases, 23 fail to execute entirely ($CMS=0$), while the remaining failures cluster between CMS values of approximately 0.5 and 0.95, indicating that many are near-correct simulations rather than complete failures. To determine whether the remaining performance gap arises from insufficient reasoning rather than insufficient knowledge, we repeated the \textbf{+Cat+PDF} evaluation using higher reasoning-effort settings. Increasing GPT-5.4 reasoning from medium to high and xhigh improves success only marginally, from 49 to 52 tasks. In contrast, adding structured domain knowledge increases performance from 0 to 49 tasks. These results indicate that access to the appropriate information is more important than additional reasoning depth.

\subsection{\textbf{Agentic execution and iterative repair}}
\label{sec:results-agentic}
The final two configurations augment knowledge injection with execution-driven feedback and iterative repair. The generic \textbf{Codex CLI} agent achieves the highest overall performance, successfully solving 63 of 86 Reproduce tasks (73\%). Our \textbf{Custom Multi-Agent} system solves 57 of 86 tasks (66\%). Both substantially outperform the strongest single-shot configuration (Fig. ~\ref{fig:headline}). Importantly, both agentic methods use the same PHITS manual and structured catalog available to the \textbf{+Cat+PDF} baseline. The improvement therefore arises primarily from the ability to execute PHITS, inspect outputs, diagnose failures, and revise candidate solutions.

The performance difference between the two agentic approaches should not be interpreted as a definitive ranking of architectures.  \textbf{Codex CLI} employs unrestricted iteration with xhigh reasoning effort, whereas our  \textbf{Custom Multi-Agent} framework uses bounded repair budgets and selective allocation of reasoning resources. A detailed inspection shows that the \textbf{Repairer} in our agent issued 206 \texttt{file\_search} queries across 32 repair sessions, with 67\,\% of those queries targeting tally syntax, which is the most error-prone PHITS feature in our corpus. This provides evidence that the agent is dedicating resources to address the performance bottleneck rather than rephrasing existing context. The comparison is therefore best viewed as an operational tradeoff between performance and computational cost. 

This distinction becomes apparent when inference cost is considered. \textbf{Codex CLI} requires approximately \$0.82 per task, whereas the \textbf{Custom Multi-Agent} architecture requires approximately \$0.13 per task. Our multi-agent system employs a bounded repair budget of five iterations together with a tiered reasoning strategy, using medium reasoning for the  \textbf{Manager} and \textbf{Composer} agents and \texttt{xhigh} reasoning only for the  \textbf{Repairer} when needed. As a result, it achieves a substantially lower per-task token cost than Codex CLI, which operates with an unrestricted \texttt{xhigh}-reasoning loop. This reduction in computational cost makes the proposed architecture a more efficient deployment strategy at comparable performance levels. 

Overall, these results suggest that execution feedback provides an additional 9–16 percentage point improvement beyond structured knowledge injection alone. Knowledge remains the dominant contributor to performance, but agentic execution enables recovery from many errors that cannot be resolved within a single inference pass.

\subsection{Failure mode analysis}
\label{sec:failures}

\begin{figure}
\centering
\includegraphics[width=0.9\linewidth]{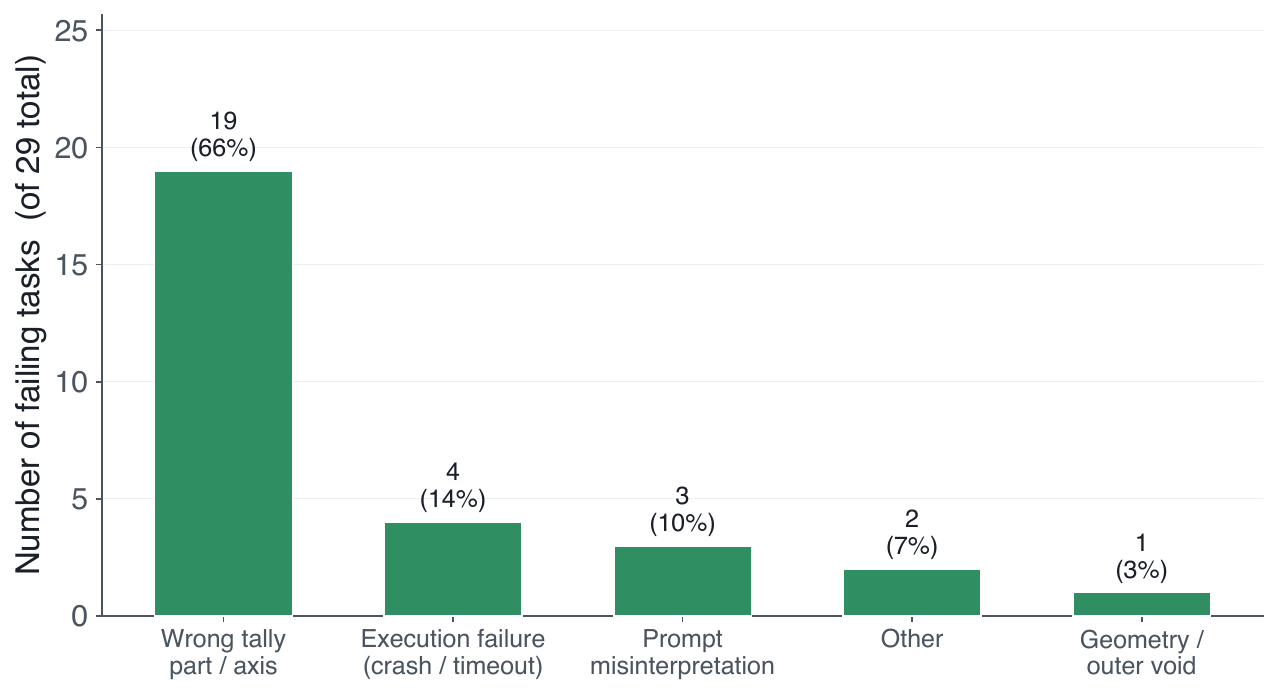}
\caption{Dominant root causes of the 29 Reproduce-track failures of the \textbf{Custom Multi-Agent} system ($\CMS < 0.95$). Most failures (25 of 29) occur after successful PHITS execution ($\EX = 1$) and therefore reflect incorrect selection or configuration of physical observables rather than syntax generation errors. Only four failures arise from execution errors ($\EX = 0$). Incorrect tally configuration is the dominant residual failure mode, highlighting the importance of domain judgment in radiation-transport modeling.}
\label{fig:fail-cats}
\end{figure}

To understand the remaining limitations of AI-assisted PHITS generation, we examined the 29 \emph{Reproduce} tasks that our \textbf{Custom Multi-Agent} system failed to solve at the CMS $\geq 0.95$ threshold. Figure ~\ref{fig:fail-cats} summarizes the dominant failure categories. Twenty-five of the 29 failures successfully execute in PHITS ($EX=1$), indicating that the generated input files are syntactically valid. Only four failures arise from execution errors. The majority of remaining failures therefore originate from deficiencies in physical fidelity rather than parser correctness. The dominant failure mode is incorrect tally configuration, which accounts for 19 of 29 failures (66\%). These cases typically involve incorrect particle selections, output units, scoring geometries, or energy-bin definitions. Execution failures account for four tasks (14\%), prompt misinterpretation for three tasks (10\%), and geometry-related errors for only a single task. \textbf{Codex-CLI} failures reveal the same qualitative pattern.

Three representative examples illustrate the pattern of the failure stemming from the lack of domain knowledge. In the first case, the prompt requested a current spectrum in units of $1/\mathrm{cm^2/MeV/source}$. The agent correctly selected the \texttt{[T-Cross]} tally and generated a valid input deck, but specified \texttt{unit = 1} ($1/\mathrm{cm^2/source}$) instead of \texttt{unit = 2}, omitting the required per-MeV normalization. In the second case, the prompt requested scoring of ionization and excitation events. The agent generated a syntactically valid \texttt{[T-Interact]} tally but omitted the \texttt{output} keyword, causing PHITS to revert to its default observable, namely nuclear-reaction density. In the third example, the agent generated a valid LET spectrum but selected \texttt{unit = 6} (dose) rather than the requested \texttt{unit = 14}, corresponding to the microdosimetric quantity $L,d(L)$.  In all three cases, the failure arose from selecting an incorrect physical quantity, normalization, or output representation. Such examples are largely absent from the current catalog, which functions primarily as a keyword and constraint reference. We therefore hypothesize that future gains may require curated training examples or post-training datasets that capture expert choices of observables and tally configurations rather than additional parser documentation alone.

A second category of failures consists of agent-specific regressions. Four of the 29 failed tasks correspond to cases that are successfully solved by the strongest single-shot configuration (\textbf{+Cat+PDF}) but are subsequently broken by the agentic repair process. Inspection of the \textbf{Repairer} traces indicates that these failures occur when the repair process modifies portions of an already correct specification while attempting to address an unrelated warning or static-checking issue. This behavior appears to reflect a limitation of the repair policy rather than a limitation of the underlying model and could likely be mitigated through stricter edit constraints or localized repair strategies.

These results indicate that current models are increasingly capable of generating syntactically valid PHITS simulations. The remaining challenge lies in incorrect selection or configuration of physical observables, suggesting limitations in \emph{domain judgment} rather than code generation.

\subsection{Independent physics validation}
\label{sec:textbook}
The CMS metric evaluates agreement between generated and reference simulations within \PHITSBench{}. A natural question, however, is whether high benchmark scores correspond to physically meaningful radiation transport calculations rather than successful reproduction of benchmark-specific outputs. To address this question, we performed an independent validation study using four canonical radiation-transport problems that do not appear anywhere in \PHITSBench{}. These benchmark problems were selected because they probe distinct aspects of particle transport physics and are commonly used for validation of radiation transport codes. Specifically, we considered: (i) the proton Bragg peak in water, which tests charged-particle stopping power and energy deposition; (ii) narrow-beam gamma attenuation in lead, which tests photon interaction cross sections and shielding calculations; (iii) the practical range of fast electrons in water, which probes electron transport and energy-loss processes; and (iv) the full-energy photopeak generated by a Cs-137 gamma source in a NaI(Tl) scintillation detector, which tests photon transport and detector response.

\begin{figure}[tp]
\centering
\includegraphics[width=\linewidth]{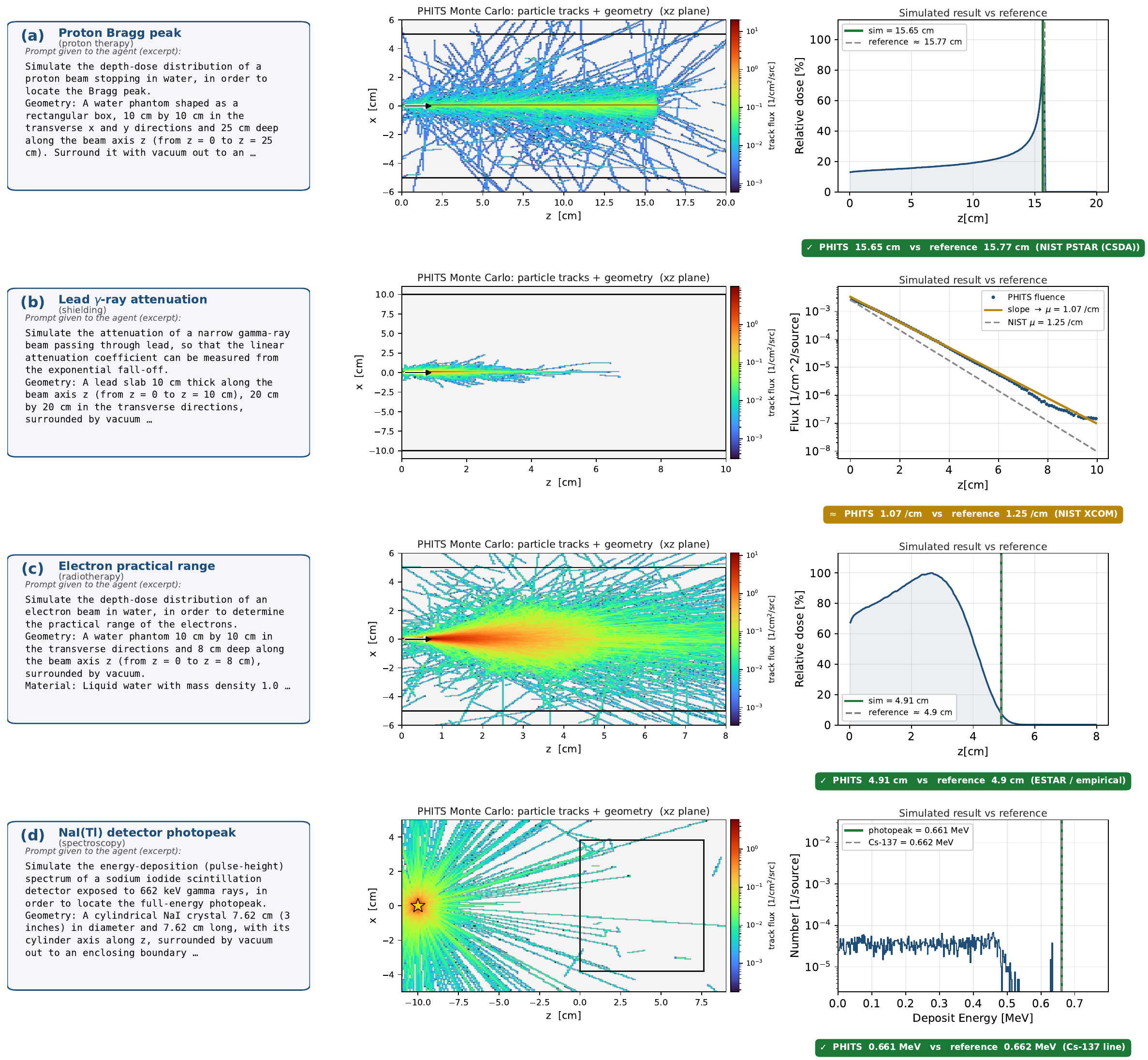}
\caption{Independent validation of AI-generated PHITS simulations using four canonical radiation-transport benchmarks not included in \PHITSBench. The left panel shows the natural language prompts. The center panels show the simulated particle tracks and geometry, while the right panels compare the resulting observables with reference values. Green (a,c,d) indicates agreement within Monte Carlo statistical uncertainty (relative errors of  about 0.2\,\% for the proton Bragg peak dose, 0.4\,\% for the electron depth-dose, 1.5\,\% for the NaI photopeak, and 0.1\,\% for the lead fluence). Amber (b) denotes a physically explained discrepancy arising from tally selection rather than an error in the underlying transport calculation.}
\label{fig:textbook}
\end{figure}

The proton benchmark evaluates whether the model can correctly reproduce the characteristic Bragg peak observed when energetic charged particles slow down in matter (Fig.~\ref{fig:textbook}a). As protons traverse water, they gradually lose kinetic energy through ionization and excitation of the medium. Because the stopping power increases as the particle slows, most energy is deposited near the end of the proton's range, producing a pronounced peak in deposited dose. Accurate prediction of the Bragg peak location is essential in proton therapy and charged-particle shielding applications.

The gamma attenuation benchmark evaluates the exponential reduction of photon intensity as gamma rays traverse an absorbing material (Fig.~\ref{fig:textbook}b). In this case, a narrow beam of photons passes through lead, where interactions such as photoelectric absorption, Compton scattering, and pair production progressively remove photons from the primary beam. The measured attenuation coefficient therefore provides a sensitive test of photon interaction physics and shielding calculations.

The electron-range benchmark evaluates transport of energetic electrons in water (Fig.~\ref{fig:textbook}c). Unlike heavy charged particles, electrons experience significant multiple scattering and stochastic energy-loss processes, resulting in broader energy deposition profiles and more diffuse stopping behavior. Accurate prediction of electron penetration depth is important for radiation dosimetry and detector modeling.

The detector benchmark evaluates generation of the characteristic Cs-137 full-energy photopeak in a NaI(Tl) scintillation detector (Fig.~\ref{fig:textbook}d). Cs-137 emits a 662-keV gamma ray that deposits its full energy within the detector when all secondary interactions are contained. Reproducing the photopeak position therefore requires correct modeling of photon interactions, energy deposition, and detector geometry.

For each benchmark, \textbf{Codex CLI} received only a natural-language description of the physical problem together with the PHITS manual and structured knowledge catalog (\textbf{+Cat+PDF}). No benchmark-specific templates or examples were provided. The resulting PHITS input files were executed in PHITS 3.34, and the principal observables were compared against established reference data (Fig.~\ref{fig:textbook}), including NIST PSTAR and ESTAR stopping-power tables for charged particles~\citep{nist_star_2017}, the NIST XCOM photon cross-section database~\citep{nist_xcom_2010}, and evaluated nuclear decay data for Cs-137~\citep{browne_2007_a137}.

Figure~\ref{fig:textbook} shows that three of the four generated simulations reproduce the corresponding reference observables within Monte Carlo statistical uncertainty. The simulated proton Bragg peak occurs at a depth of 15.65 cm, compared with the NIST CSDA range of 15.77 cm. This small offset is physically expected because the maximum deposited dose occurs slightly before the particle reaches its stopping point. The electron practical range (4.91 cm versus 4.90 cm) and the NaI(Tl) photopeak energy (0.661 MeV versus 0.662 MeV) agree with reference values within simulation uncertainty. The remaining benchmark, photon attenuation in lead, produced an attenuation coefficient approximately 14\% lower than the reference value (1.07 versus 1.25 cm$^{-1}$). Examination of the generated input revealed that the attenuation coefficient was extracted from a track-length tally that includes scattered photons in addition to uncollided primary photons. Consequently, the discrepancy reflects a modeling choice rather than an error in the underlying transport calculation and could be eliminated by replacing the tally with a detector-fluence or first-collision scoring method. These results suggest that successful benchmark performance corresponds to physically meaningful simulations and that the generated inputs reproduce a diverse set of canonical radiation-transport phenomena spanning charged-particle stopping, photon attenuation, electron transport, and detector response.

\section{Discussion}
\label{sec:discussion}

The central finding of this study is that AI-assisted generation of radiation transport simulations is already feasible for a substantial fraction of practical PHITS workflows, provided that domain knowledge is supplied in a structured and accessible form. Across the \emph{Reproduce} benchmark, the largest performance gain arises not from increased reasoning effort or more sophisticated agent architectures, but from the introduction of a structured PHITS knowledge catalog (Fig.~\ref{fig:headline}). Agentic execution provides additional improvements, but its contribution is secondary to the availability of machine-readable domain knowledge. This observation has implications both for radiation transport modeling specifically and for AI-assisted scientific computing more broadly. The results suggest that many current limitations of scientific AI systems arise not because foundation models lack general reasoning ability, but because critical domain knowledge remains embedded in documentation, source code, and expert practice in forms that are difficult for AI systems to access and utilize effectively.

\subsection{Implications for radiation transport modeling}

The benchmark tasks examined in this work closely mirror the workflows routinely performed by PHITS users during simulation development, validation, and deployment. Consequently, the observed performance provides insight into where AI assistance may become useful in practical radiation transport studies.

The strongest near-term opportunity is simulation quality assurance. The \emph{Edit} and \emph{Repair} tracks, which approach saturation under single-shot prompting, correspond to common debugging and validation tasks such as identifying missing parameters, correcting malformed geometry definitions, repairing invalid tally configurations, or detecting inconsistencies between source and scoring specifications. These activities often consume substantial time during model development, particularly because many PHITS inputs are syntactically valid yet physically incorrect. The ability of an AI assistant to automatically flag such issues before long production runs could reduce both development time and the risk of propagating incorrect simulations through larger engineering studies.

The benchmark also suggests a possible role for AI-assisted verification of the radiation-transport workflows. Because PHITSBench evaluates both execution success and agreement with known transport observables, it provides a framework for distinguishing between simulations that merely run and simulations that produce physically meaningful results. Such execution-grounded validation could be particularly valuable in shielding and facility design ~\citep{iwamoto2022accel,takada2018proton}, detector development and medical physics ~\citep{satoh2022detector}, reactor analysis~\citep{cruz2024reactor}, and space and planetary
science~\citep{sato2011space}, where small configuration errors can lead to significant discrepancies in predicted particle fluxes, dose distributions, or detector responses.

Finally, PHITSBench may serve as a useful educational resource. Each benchmark task pairs a natural-language physics problem with a validated reference simulation and quantitative scoring framework.  Such examples could support training of students, researchers, and designers, and help bridge the gap between understanding transport physics and implementing that understanding in executable simulation workflows.

\subsection{Community infrastructure opportunities}

The strongest performance improvements observed in this work originate from resources that are largely external to the foundation model itself. This suggests several opportunities for community-driven development. 

The first opportunity is the creation of a machine-readable PHITS knowledge base.  The single-shot \textbf{+Cat+PDF} configuration produces the largest cumulative improvement over the no-knowledge baseline in the entire ablation study (Fig.~\ref{fig:headline}, from 0\% to 57\,\%). The underlying catalog consists of 11 Markdown files totaling approximately 7,300 lines of structured content. It was generated through LLM-assisted parsing of the PHITS Fortran source code and then manually audited, refined, and organized by a domain expert. This catalog serves as a proof of concept, which can be built upon to create a structured machine-readable corpus (in the YAML or JSON format), updated in lock-step with new PHITS version releases.  It should capture keyword definitions, accepted parameter ranges, default values, parser constraints, compatibility rules, and common error patterns in a format suitable for both human users and AI systems. Because much of this information already exists within the PHITS source code and documentation, such a resource would be relatively inexpensive to maintain while providing substantial value.

A second opportunity is the development of curated domain-training datasets. The remaining benchmark failures frequently involve choices that experienced users make implicitly when selecting observables, scoring methods, or tally configurations. These decisions are rarely documented formally and, therefore, are difficult for AI systems to learn through retrieval alone. Even a few thousand such expert-curated examples, collected through a community submission process and reviewed by experienced PHITS users, could provide a valuable foundation for domain-specific post-training, including LoRA (low-rank adaptation), full fine-tuning, and DPO (direct preference optimization). Such a dataset could help move the model beyond generic code generation and retrieval-augmented assistance toward behavior more closely resembling that of an experienced PHITS practitioner.

A third, potentially high-impact direction is the development of a verifiable reinforcement-learning environment. One of the most distinctive properties of \PHITSBench{} as well as other Monte Carlo simulation tasks is that performance can be evaluated using a deterministic, computable reward. Given a candidate input file and a reference simulation, the resulting CMS is a quantitative measure of correctness rather than a subjective judgment. This is an excellent setting in which reinforcement learning from verifiable environments can be applied effectively. A natural next step would be to package \PHITSBench{} as a Gymnasium-compatible environment, expose CMS as the reward signal, and enable models to improve through GRPO (Group Relative Policy Optimization) or related reinforcement-learning algorithms. 

Significant engineering challenges remain, including ensuring reward stability, maintaining sufficient task diversity, preventing benchmark overfitting, and managing the licensing and computational costs associated with large-scale training. Although these directions are beyond the scope of this work, we believe they represent one of the most promising paths toward transforming AI-assisted MC transport workflows from useful code-generation tools into reliable, human-supervised collaborators. Such a benchmark could become a shared community resource whose value increases as multiple groups contribute data, methods, and models within a common evaluation framework.

\subsection{Limitations}
\label{sec:limitations}

Several limitations should be considered when interpreting these results. First, the study evaluates a single model family (GPT-5.4) and a limited set of agent architectures. Future models may exhibit different behavior, particularly with respect to knowledge utilization and self-correction. Second, the benchmark focuses exclusively on PHITS version 3.34. Although the methodology is broadly applicable, changes in syntax or functionality across future PHITS releases may affect absolute performance values. Third, the knowledge catalog itself is a manually curated artifact. While this reflects realistic domain-engineering practice, it introduces a dependence on catalog quality that has not been systematically studied. Future work should investigate how benchmark performance varies as a function of knowledge representation, completeness, and organization. Finally, the benchmark evaluates simulation generation rather than scientific problem solving in the broader sense. Producing a valid PHITS input file is only one step within a larger engineering workflow that includes model verification, uncertainty quantification, experimental validation, and interpretation of results.

\section{Conclusions}
\label{sec:conclusion}

We introduced \PHITSBench, an execution-scored benchmark for evaluating AI-assisted generation of PHITS radiation-transport simulations from natural-language descriptions, and used it to evaluate five GPT-5.4-based configurations ranging from zero-shot prompting to knowledge-augmented and agentic workflows. The results reveal a clear hierarchy of factors limiting performance.

Specifically, we find that (i) structured knowledge injection provides the largest performance gain on the most challenging benchmark track, increasing Reproduce-task success from 0\% to 57\%; (ii) agentic execution yields a further, though smaller, improvement to 66-73\%, depending on the architecture; and (iii) the remaining failures can be traced to a small number of identifiable root causes that point as much to gaps in domain-specific infrastructure as to limitations of the underlying model. 

We conclude that for highly specialized scientific simulation environments such as PHITS access to structured domain knowledge is a more important determinant of performance than additional reasoning effort alone. We therefore view machine-readable knowledge bases, curated domain-training datasets, and verifiable reinforcement-learning environments as promising directions for future development for scientific simulation ecosystems beyond PHITS.

\section*{Acknowledgement}
\addcontentsline{toc}{section}{Acknowledgement}
The authors thank Tongge Yu (MIT) for contributing to the conception and early brainstorming of this project. We are grateful to Dr. Tatsuhiko Sato of the PHITS development team for insightful discussions and constructive feedback. We also thank Dr. Duo Xu and Charles Lai (MIT) for their contributions to the design and evaluation of the benchmark. OpenAI API credits supporting this work were provided by OpenAI through the MIT AI Hackathon.
\section*{Disclosure statement}
\addcontentsline{toc}{section}{Disclosure statement}
No potential conflict of interest was reported by the authors.


\bibliography{references}

\end{document}